Book: Soft Robotics in Biomedical Sciences: Current Status and Advancements (Eds: Deepak Gupta, Rishabha Malviya, Rekha Raja, Sonali Sundram

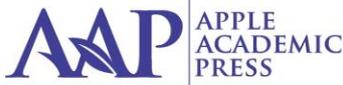

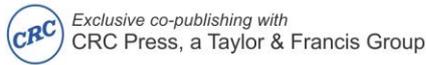

**BOOK TITLE:** Soft Robotics in Biomedical Sciences: Current Status and Advancements
**BOOK EDITORS:** Deepak Gupta, Rishabha Malviya, Rekha Raja, Sonali Sundram
**CHAPTER TITLE:** Bio-inspired Soft Grippers for Biological Applications
**CHAPTER AUTHORS:** Rekha Raja, Ali Leylavi Shoushtari

# Contents

## Chapter 8

**Bio-inspired Soft Grippers for Biological Applications**

*Rekha Raja[1*],  Ali Leylavi Shoushtari[2]*

[1]*Dept. of Artificial Intelligence, Radboud University, Nijmegen, Netherlands (*Corresponding author: rekha.cob@gmail.com )*

[2]*Wageningen Robotics, Wageningen, Netherlands (Email: ali.leylavishoushtari@wagrobotics.com)*

**Abstract**: The field of bio-inspired soft grippers has emerged as a transformative area of research with profound implications for biomedical applications. This book chapter provides a comprehensive overview of the principles, developments, challenges, and prospects of soft grippers that draw inspiration from biological systems. Bio-inspired soft grippers have gained prominence due to their unique characteristics, including compliance, adaptability, and biocompatibility. They have revolutionized the way we approach biomedical tasks, offering safer interactions with delicate tissues and enabling complex operations that were once inconceivable with rigid tools.

The chapter delves into the fundamental importance of soft grippers in biomedical contexts. It outlines their significance in surgeries, diagnostics, tissue engineering, and various medical interventions. Soft grippers have the capacity to mimic the intricate movements of biological organisms, facilitating minimally invasive procedures and enhancing patient outcomes. A historical perspective traces the evolution of soft grippers in biomedical research, highlighting key milestones and breakthroughs. From early attempts to emulate the dexterity of octopus tentacles to the latest advancements in soft lithography and biomaterials, the journey has been marked by ingenuity and collaboration across multiple disciplines.

Motivations for adopting soft grippers in biomedical applications are explored, emphasizing their ability to reduce invasiveness, increase precision, and provide adaptability to complex anatomical structures. The requirements and challenges in designing grippers fit for medical contexts are outlined, encompassing biocompatibility, sterilization, control, and integration. The chapter also sheds light on the state-of-the-art literature in bio-inspired soft grippers, showcasing the cutting-edge research that is propelling this field forward. Examples of soft grippers inspired by nature's design principles are presented, illustrating the wide array of possibilities and applications.

## 8.1 INTRODUCTION

Soft grippers are devices that are designed to grasp and manipulate objects without causing damage. They are typically made of soft, flexible materials such as silicone, rubber, or hydrogels, and are capable of conforming to the shape of the object being gripped. Bio-inspired soft grippers are designed based on principles found in nature, such as the way that an octopus uses its tentacles to grasp objects or the way an elephant uses its trunk to grasp, pick up and release objects without breaking them as shown in Figure 8.1.

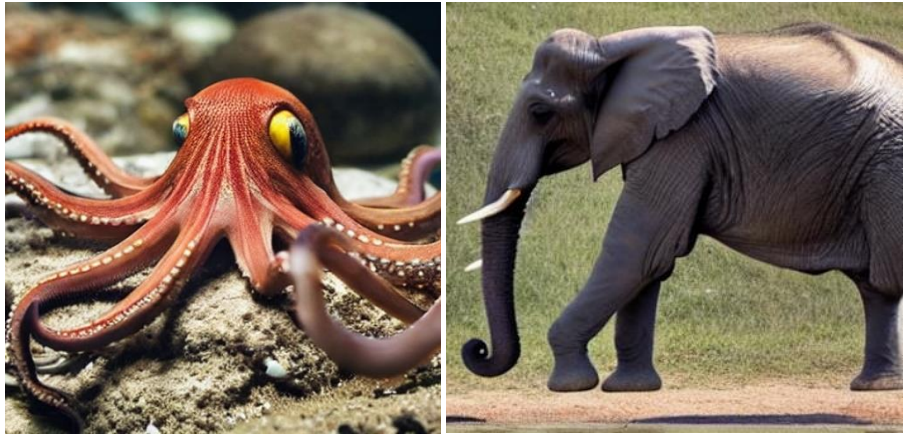

**Figure 8.1:** Octopus using its tentacles to grasp objects (left), elephant using its trunk to grasp objects.

One of the main advantages of a soft gripper is its adaptability, while conventional industrial grippers are known for their speed, precision, and power. They have been widely adopted in many industries, including manufacturing, automotive, and electronics, where their rigid structures and high performance can deliver impressive results. However, in applications that require gentle manipulation and handling of delicate objects, such as food processing, medical procedures, or rehabilitation, conventional grippers can lack in reachability, adaptability, degree of freedom, and safe interaction that can pose a significant risk of damaging the manipulated objects or patients. Moreover, they are often unsuitable for performing tasks in unstructured or dynamic environments due to their fixed and rigid structure. Unlike conventional grippers that are designed to operate in a controlled environment, soft grippers can handle objects in complex and dynamic environments. This opens up new opportunities for applications in areas such as healthcare, agriculture, and manufacturing, where soft robots can perform tasks that were previously impossible or unsafe.

Biomedical applications exhibit specific features that make soft grippers a suitable choice for handling objects in these contexts. These features include: A- Safety requirements of grasping and handling objects: Soft grippers offer a compliant and gentle touch, reducing the risk of damage to delicate biological samples. This is crucial in fields like tissue engineering and regenerative medicine, where precision and control are necessary [1]. B- Safety requirements during maneuvering in the vicinity of organs/tissues: Soft grippers' flexibility and adaptability enable them to navigate complex anatomical structures safely. They can conform to the irregular shapes of organs or tissues, reducing the risk of injury during procedures[2]. C- Unpredictability of the shape of objects: Soft grippers can adapt to objects with unknown or varying shapes. Their compliance allows for effective grasping and manipulation, regardless of the object's specific geometry [3]. D- High deformability of objects/organs: Soft grippers excel in handling deformable objects and organs. Their compliant nature enables them to grasp and manipulate soft and flexible structures, such as organs or tissues, without causing damage [4]. Soft grippers possess exceptional capabilities that make them well-equipped to perform the tasks required in biomedical applications. Their ability to conform to objects with unknown or irregular shapes allows for secure and gentle grasping, minimizing the risk of damage or injury. The flexibility and adaptability of soft grippers enable them to handle deformable objects, such as soft tissues, with care and precision. The non-stick surfaces or coatings of soft grippers reduce the risk of contamination, making them suitable for maintaining sterile conditions in applications like in vitro fertilization and cell culture. Additionally, soft grippers can provide the necessary accuracy and precision required for tasks such as targeted drug delivery or delicate surgical procedures. Their cost-effectiveness further enhances their attractiveness as a solution for biomedical applications. Soft grippers find relevance in various biomedical applications, some of which already have dedicated soft robotic platforms developed. These applications include: A- Robotic surgery: soft grippers can assist in delicate surgical procedures, providing controlled and precise manipulation of tissues or organs. This can enhance surgical outcomes and minimize the invasiveness of procedures [5]. B- Endoscopy: Soft grippers are valuable in endoscopic procedures, where flexibility and adaptability are essential for handling flexible and deformable objects, such as endoscope catheters or tissue samples [6]. C- Tissue engineering and regenerative medicine: soft grippers enable the delicate handling and manipulation of living cells and tissues, facilitating processes like tissue assembly, transplantation, or microsurgery [7]. D- Rehabilitation robotics: soft grippers can aid in the manipulation of prosthetic devices or assist individuals with impaired motor function. They offer gentle interaction and improved control for rehabilitation purposes [8]. Please note that the provided references cover the general concepts and applications of soft robotics in biomedical contexts and may not explicitly focus on soft grippers. However, they provide valuable insights into the broader field of soft robotics in biomedical applications.

To develop a soft gripper for healthcare application must meet several requirements to be effective and safe. For example, soft grippers must be biocompatible, meaning that they do not cause a harmful immune response when in contact with living tissue. This is particularly important in applications where the gripper will be in direct contact with biological samples or implanted within the body. It must be designed and fabricated to ensure sterility, non-toxicity, flexibility, durability, and easy to use. Overall, the requirements for soft grippers used in biomedical applications are focused on ensuring safety, efficacy, and ease of use in handling delicate biological samples or performing surgical procedures. By meeting these requirements, soft grippers can be powerful tools in the fields of tissue engineering, regenerative medicine, drug delivery, and surgery.

While soft grippers have several advantages over conventional rigid grippers in biomedical applications, they also present unique challenges. For example, while selecting materials certain parameters must be taken into consideration such as There are some of the key challenges associated with using soft grippers in biomedical applications such as, the material must be biocompatible, non-toxic, and able to withstand repeated use and sterilization without degradation or loss of functionality. Soft grippers are focused on improving their functionality, durability, and ease of use while maintaining safety and biocompatibility. Addressing these challenges will be essential for unlocking the full potential of soft grippers in fields such as regenerative medicine, drug delivery, and surgery.

Despite huge promises and also intensive improvement of soft robotic grippers they still have a long way to reach the market. Researchers showed that there are several criterias for a field e.g., agrofood robotics needed to be met for a robotic gripper to be called "commercially ready" for market. Biomedical applications are not an exception to this principle, and therefore further development is necessary to address the technological and scientific challenges specific to this field. One crucial aspect is improving the dexterity and control of soft grippers to enable more intricate and precise manipulation of delicate biological samples. Enhancing their sensing capabilities to provide feedback and real-time information during gripping and handling is also essential for optimal performance in biomedical tasks. Furthermore, the integration of soft grippers with advanced imaging and sensing technologies can enable better understanding and visualization of the objects being manipulated. This integration will facilitate more accurate and efficient procedures, particularly in complex surgical interventions or tissue engineering processes. Additionally, the development of reliable and robust soft materials with enhanced durability and biocompatibility is crucial to ensure the long-term functionality and safety of soft grippers in biomedical applications. Overall, continued research and innovation are necessary to further refine and advance soft gripper technology, enabling them to effectively meet the unique challenges and requirements of biomedical applications. By addressing these challenges, soft grippers have the potential to revolutionize various aspects of healthcare and contribute to advancements in fields such as surgery, diagnostics, and regenerative medicine.

Bio-inspired soft grippers draw inspiration from nature, particularly the characteristics and functionalities of organisms or biological structures, to enhance the capabilities of robots in achieving the goals of biomedical applications. By mimicking biological systems, these grippers offer unique advantages that can address the challenges specific to the biomedical field. Here are a few ways bio-inspired soft grippers can assist robots in achieving these goals:

**Adaptability and Compliance:** Bio-inspired soft grippers replicate the compliance and adaptability found in natural organisms. By incorporating flexible materials and designs that mimic biological structures such as tentacles or soft muscles, these grippers can conform to the shape of objects and handle high deformability. This adaptability enables gentle handling of delicate biological samples without causing damage.

**Sensing and Feedback:** Many biological organisms possess sophisticated sensory systems, allowing them to perceive and respond to their environment. Bio-inspired soft grippers can integrate sensors or sensory feedback mechanisms to enhance their perception capabilities. This enables robots to receive real-time information about the objects being manipulated, providing crucial feedback for precise control and manipulation in biomedical tasks.

**Enhanced Biocompatibility:** Bioinspiration can guide the development of soft grippers with enhanced biocompatibility. By studying and replicating the structures and materials found in biological organisms, soft grippers can be designed to interact seamlessly with biological systems. This includes selecting materials that are biologically inert, non-toxic, and compatible with the surrounding tissues or cells. Improved biocompatibility ensures that soft grippers can be safely used in contact with biological samples without causing adverse reactions or complications.

**Self-Healing and Self-Repair:** Certain bio-inspired soft grippers take inspiration from self-healing mechanisms found in nature. By incorporating materials that possess self-healing properties, these grippers can autonomously repair any damage or wear, ensuring their long-term functionality and reducing the need for maintenance or replacement. Energy Efficiency: Many biological organisms have evolved to be energy-efficient, utilizing minimal resources for maximum performance. Bio-inspired soft grippers can incorporate energy-saving mechanisms, such as passive actuation or energy recovery systems, to improve their efficiency and prolong their operating time. This can be beneficial in prolonged biomedical procedures or tasks requiring prolonged gripping or manipulation. By incorporating bio-inspired design principles into soft gripper technology, robots can achieve a higher level of functionality, adaptability, and safety in biomedical applications. These grippers enable precise manipulation of delicate samples, provide real-time feedback, navigate complex environments, and optimize energy usage. Ultimately, bio-inspired soft grippers contribute to the development of advanced robotic systems that closely resemble and emulate the capabilities of natural organisms, opening up new possibilities for advancements in the biomedical field.

In this chapter, we will explore the field of bio-inspired soft grippers in-depth, focusing on the design principles, materials, and technologies that underpin this exciting new field. We will examine the advantages and limitations of bio-inspired soft grippers, compare them to the other soft grippers, conventional industrial grippers, and discuss the challenges and opportunities for future research and development. This chapter will provide readers with a comprehensive understanding of bio-inspired soft grippers and its potential to transform the way we interact with the world around us.

## 8.2 BACKGROUND

For centuries, engineers have been inspired by natural organisms, driving them to create increasingly capable machines for various applications [9-10]. The influence of biological systems on the development of soft, flexible robots is particularly fascinating. By observing how organisms distribute stress throughout their bodies and adapt to rapidly changing environments, engineers have gained valuable insights that have significantly improved the design of soft-bodied machines. In comparison to their rigid-body counterparts, these soft robots offer unique advantages in terms of their softness and body compliance [11].

One of the key benefits of soft robots is their inherent safety when interacting with humans, which is in stark contrast to most rigid-bodies robots. This makes them highly desirable for human-in-the-loop cooperative operations. Additionally, soft robots possess exceptional adaptability, compliance, and flexibility [12]. They have the remarkable ability to continuously deform with a high degree of freedom (DOFs) [9]. This flexibility enables them to navigate complex and unstructured environments [13], manipulate delicate objects, and interact with living organisms in a gentle and non-intrusive manner.

Nature can offer valuable insights for designing new versatile grippers that can outperform current robotic grippers. Researchers have studied various biological grippers, including manual grippers like the human hand, spinal grippers like a snake's body, and muscular hydrostats like an octopus arm [14]. These biological grippers exhibit a high level of versatility in grasping objects of different sizes, shapes, and materials. One key lesson from nature is the combination of gripping mechanisms used by biological grippers, namely mechanical interlocking, friction, and adhesion [14]. Mechanical interlocking involves wrapping a gripper around an object and manipulating it via normal compressive forces. Friction grip generates parallel shearing force between the gripper and an object of similar size. Adhesion, characterized by normal tensile force, allows animals like tree frogs and geckos to grip on flat surfaces. To mimic these principles, researchers have developed bioinspired robotic grippers. For example, a hybrid gecko-inspired finger-like gripper combines both friction and adhesion to grasp objects with complex geometries [15]. This approach allows for the regulation of mechanical contact stresses and minimizes the risk of crop damage [20]. Additionally, advancements in soft robotics have led to the design of inflatable origami-based actuators, which are ultra-lightweight and inspired by plant structures [16]. These actuators can be integrated into grippers, providing them with flexibility and adaptability, akin to biological grippers. However, there are still challenges in creating bioinspired grippers with enhanced versatility. Integrating stiff and soft bodyware, actuators, sensors, and functional surfaces is crucial and requires advances in robotic technologies and material science, especially in the field of soft robotics [3].

Biohybrid approaches, combining organic and synthetic components, may complement soft robotic technologies [17]. By drawing on the architecture and functioning of biological grippers and fusing them into smart, hybrid, and adaptable grippers, researchers can close the gap in versatility between biological and agrorobotic grippers. This approach will enable the development of grippers capable of autonomously operating in complex environments, firmly yet gently gripping crops with unknown properties, and effectively assisting in handling complex and delicate objects, especially in the field of agrorobotics and beyond.

More recently, there has been a growing interest in the development of soft grippers for bio-medical applications, driven by the need for devices that can handle biological samples without causing damage. In the early 2000s, researchers began exploring the use of soft grippers in bio-medical applications such as tissue engineering and regenerative medicine. In 2003, a research group led by Robert Langer at the Massachusetts Institute of Technology (MIT) developed a soft robotic hand capable of grasping and manipulating delicate objects, including living cells and tissues. The Robotic hands could be count as the first soft bio-inspired grippers used in biomedical applications. For instance Smart Hand from Scuola Superiore Sant'Anna (Figure 8.2) is one of the pioneer work used as cyberhand for amputees [18].

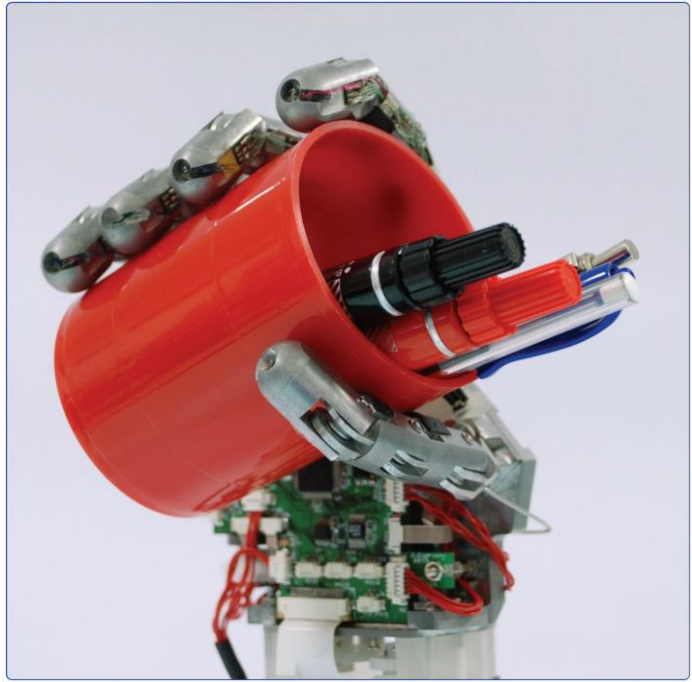

Figure 8.2 Human hand as one of the first sources of inspirations for robotics grippers in biomedical applications [18].

Since then, there has been a significant increase in research on bio-inspired soft grippers for bio-medical applications. Researchers have explored various materials and fabrication techniques for soft grippers, as well as different design principles inspired by nature. For example, some soft grippers are designed based on the way that octopuses use their tentacles to grasp objects, while others are inspired by the way that geckos use their feet to climb vertical surfaces.

Nature provides a wealth of examples showcasing the effectiveness of soft, mobile structures, which can serve as inspiration for creating man-made structures. One such example is the muscular hydrostat, a biological structure capable of bending, twisting, and extending with ease. Invertebrates like squids, jellyfish, and giant earthworms exhibit remarkable body deformations that allow them to adapt to their surroundings and access food and shelter more efficiently. Among these creatures, octopuses are particularly fascinating to researchers due to their rapid limb extension, especially when striking their prey. By gaining a deep understanding of the morphology and structures of these arthropods and invertebrates, developers can create innovative and resilient designs for soft robots [19].

## 8.3 BIOLOGICAL INSPIRATION FOR SOFT GRIPPER DESIGN IN BIOMEDICAL APPLICATIONS

Bio-inspired soft robotics grippers encompass a wide range of designs and mechanisms, drawing inspiration from various biological systems. Here are some types of bio-inspired soft robotics grippers commonly explored in research and development:

1. **Gecko-Inspired Grippers:** These grippers mimic the adhesive properties of gecko feet. They typically incorporate micro/nanostructured surfaces or utilize dry adhesives to achieve strong and reversible adhesion. Gecko-inspired grippers can adhere to a variety of surfaces, enabling versatile gripping capabilities.

2. **Octopus-Inspired Grippers:** These grippers emulate the dexterity and flexibility of octopus tentacles. They often feature soft, elongated structures with distributed suction cups or flexible arms to achieve precise and adaptable grasping. Octopus-inspired grippers excel in manipulating objects with complex shapes and in dynamic environments.

3. **Jellyfish-Inspired Grippers:** These grippers imitate the propulsive and contracting motions of jellyfish. They usually employ soft, bell-shaped structures that rhythmically contract and expand to generate fluid flows for grasping and propulsion. Jellyfish-inspired grippers are suitable for underwater applications [20].

4. **Vine-Inspired Grippers:** These grippers draw inspiration from the climbing and wrapping behavior of vines. They utilize soft, flexible structures that can conform to and wrap around objects to achieve a secure grip. Vine-inspired grippers are often used in scenarios where grasping irregular or complex objects is required [21-22].

5. **Fish-Inspired Grippers:** These grippers mimic the undulating motion and hydrodynamics of fish tails or fins. They typically comprise flexible structures that generate fluid flows to produce propulsive forces for gripping or manipulation tasks. Fish-inspired grippers excel in underwater environments [23].

6. **Insect-Inspired Grippers:** These grippers take inspiration from the gripping mechanisms of insects, such as beetles or ants. They may utilize microstructured surfaces, hooks, or adhesive pads to achieve strong adhesion and grasping capabilities. Insect-inspired grippers are often compact and suitable for small-scale applications.

7. **Snake-inspired grippers**: It mimic the locomotion and gripping capabilities of snakes. They often feature flexible, segmented structures that can wrap around objects or conform to irregular surfaces, allowing for versatile grasping and manipulation.

8. **Caterpillar-inspired grippers:** Caterpillar-inspired grippers imitate the movement and gripping mechanisms of caterpillars. They typically employ soft, flexible structures with multiple appendages or gripping elements that can deform and conform to objects for secure grasping and controlled manipulation.

9. **Bat-inspired grippers:** Bat-inspired grippers draw inspiration from the wings and claws of bats. They may incorporate flexible, membrane-like structures or grasping elements that mimic the wing morphology and clawed fingers of bats. These grippers enable unique grasping capabilities and dexterous manipulation.

10. **Bird-inspired grippers:** Bird-inspired grippers take cues from the feet and claws of birds. They often feature articulated structures or specialized gripping elements that replicate the grasping and perching abilities of birds. These grippers are designed to securely hold objects of various sizes and shapes.

11. **Plant-inspired grippers:** Plant-inspired grippers derive inspiration from the gripping and attachment mechanisms found in certain plants. They may utilize features such as thorns, barbs, or adhesive surfaces to achieve effective grasping and adherence to objects or surfaces. Plant-inspired grippers are often used for applications that require strong and reliable gripping capabilities.

12. **Mussel-inspired grippers:** Mussel-inspired grippers mimic the adhesive properties of mussels' byssus threads. They often utilize biomimetic adhesive materials or structures that can adhere to various surfaces, allowing for secure gripping and manipulation.

13. **Crab-inspired grippers:** Crab-inspired grippers draw inspiration from the pincers or claws of crabs. They may incorporate articulated structures, rigid elements, or specialized gripping surfaces to replicate the gripping and pinching abilities of crabs. These grippers offer strong and precise grasping capabilities.

14. **Elephant trunk-inspired grippers:** Elephant trunk-inspired grippers emulate the versatile trunk of elephants. They typically feature flexible, elongated structures that can bend, twist, and conform to objects for grasping and manipulation. Elephant trunk-inspired grippers enable precise and dexterous handling of various objects.

15. **Human hand-inspired grippers:** Human hand-inspired grippers mimic the structure and functionality of the human hand. They may incorporate articulated fingers, joints, and tactile sensing elements to replicate the grasping, manipulation, and sensory capabilities of the human hand. These grippers are designed for versatile and intuitive object manipulation.

16. **Seahorse-inspired grippers:** Seahorse-inspired grippers mimic the prehensile tail of seahorses. They feature flexible, curved structures that can wrap around objects and provide a secure grip. Seahorse-inspired grippers are suitable for applications that require grasping and manipulation in aquatic environments.

**17. Spider-inspired grippers:** Spider-inspired grippers draw inspiration from the legs and adhesive capabilities of spiders. They may utilize articulated leg-like structures and adhesive materials to achieve climbing and gripping abilities similar to those of spiders. Spider-inspired grippers are often used for tasks that require climbing or adherence to surfaces.

**18. Starfish-inspired grippers:** Starfish-inspired grippers mimic the regenerative and adhesive properties of starfish arms. They can utilize soft, flexible structures and adhesive materials to achieve both grasping and attachment capabilities. Starfish-inspired grippers are designed for applications requiring gentle and adaptable gripping.

**19. Beetle-inspired grippers:** Beetle-inspired grippers take inspiration from the gripping mechanisms of beetles, such as their claws or specialized structures. They may feature robust and articulated gripping elements that replicate the strength and precision of beetle claws. Beetle-inspired grippers are suitable for applications requiring strong and secure grasping.

**20. Biomimetic tentacle grippers:** Biomimetic tentacle grippers imitate the elongated and flexible tentacles found in various marine organisms. They typically utilize soft, slender structures with active bending or suction mechanisms to achieve versatile grasping and manipulation in underwater environments. Biomimetic tentacle grippers are often used for tasks like underwater exploration and object retrieval.

Each of these bio-inspired grippers takes inspiration from specific biological systems and adapts their unique features and mechanisms to develop innovative and functional gripping solutions in the field of soft robotics.

**TABLE 8.1** List of bio-inspired grippers with their key features, applications, actuation systems, materials used, and reference papers in bio-medical applications.

| Bio-inspired grippers | Biological inspiration | Key features | General Applications | Prospective Biomedical applications | Actuation systems | Materials used | Reference papers |
|---|---|---|---|---|---|---|---|
| Gecko-inspired | Gecko feet | Adhesive micro/nanostructures | Climbing, adhesion to various surfaces | Surgical robotics | Microtubes or pneumatic systems | Adhesive micro/nanostructures, elastomers | [19, 24, 25] |
| Octopus-inspired | Octopus tentacles | Flexible, distributed suction cups | Grasping complex objects, adaptability | Minimally invasive surgery | Pneumatic or hydraulic systems | Elastomers, soft materials | [9, 26, 27] |
| Jellyfish-inspired | Jellyfish | Soft, bell-shaped structures | Underwater manipulation, propulsion | Tissue engineering scaffolds | Pneumatic or hydraulic systems | Elastomers, hydrogels | [28, 29, 30] |

| Vine-inspired | Climbing vines | Conformable, wrap around objects | Grasping irregular shapes, climbing | Endoscopic procedures | Pneumatic or tendon-driven systems | Elastomers, flexible materials | [31] |
|---|---|---|---|---|---|---|---|
| Fish-inspired | Fish tails/fins | Undulating, hydrodynamic structures | Underwater manipulation, propulsion | Underwater exploration | Hydraulic or fin-based systems | Elastomers, soft materials | [32, 33] |
| Insect-inspired | Insect gripping | Microstructured surfaces, hooks, adhesives | Precise grasping, small-scale tasks | Microsurgery, drug delivery | SMA (Shape Memory Alloy), pneumatic or tendon-driven systems | Elastomers, biomaterials | [34] |
| Snake-inspired | Snakes | Flexible, wrap-around structures | Versatile manipulation, conformability | Minimally invasive surgery | Cable-driven or pneumatic systems | Elastomers, flexible materials | [35] |
| Caterpillar-inspired | Caterpillars | Flexible, segmented gripping elements | Controlled manipulation, crawling | Soft robotics for rehabilitation | Pneumatic or tendon-driven systems | Elastomers, soft materials | [36] |

| Bat-inspired | Bats | Membrane-like structures, clawed fingers | Dexterous manipulation, grasping | Prosthetics, assistive devices | Cable-driven or pneumatic systems | Elastomers, flexible materials | [37] |
|---|---|---|---|---|---|---|---|
| Bird-inspired | Bird feet/claws | Articulated structures, specialized gripping | Secure grasping, perching | Robotic exoskeletons | Cable-driven or pneumatic systems | Elastomers, biomaterials | [38] |
| Plant-inspired | Plants | Thorns, barbs, adhesive surfaces | Strong attachment, gripping | Surgical tools, tissue engineering | Pneumatic or hydraulic systems | Elastomers, biomaterials | [39] |
| Mussel-inspired | Mussels | Biomimetic adhesives, secure attachment | Adhesion, secure gripping | Surgical tools, wound closure | Pneumatic adhesive systems | Adhesives, biomaterials | [40] |
| Crab-inspired | Crabs | Articulated structures, specialized gripping | Pinching, strong grasping | Prosthetics, rehabilitation | Pneumatic or tendon-driven systems | Elastomers, biomaterials | [41] |
| Elephant trunk-inspired | Elephant trunk | Flexible, bending, twisting structures | Delicate object manipulation | Rehabilitation robotics | Cable-driven or pneumatic systems | Elastomers, flexible materials | [42] |
| Human hand-inspired | Human hand | Articulated fingers, tactile sensing | Versatile grasping, fine manipulation | Prosthetics, rehabilitation | Cable-driven or pneumatic systems | Elastomers, biomaterials | [43] |
| Seahorse-inspired | Seahorses | Flexible, curved structures | Aquatic grasping, manipulation | Drug delivery, cell manipulation | Pneumatic or hydraulic systems | Elastomers, soft materials | [44] |
| Spider-inspired | Spiders | Articulated legs, adhesive capabilities | Climbing, surface adherence | Surgical tools, tissue engineering | Pneumatic or tendon-driven systems | Elastomers, biomaterials | [45] |
| Starfish-inspired | Starfish arms | Flexible, adhesive structures | Gentle grasping, attachment | Soft grippers, tissue engineering | Pneumatic or hydraulic systems | Elastomers, biomaterials | [46] |

| Beetle-inspired | Beetles | Robust gripping elements, specialized structures | Strong grasping, secure attachment | Prosthetics, rehabilitation | Cable-driven or pneumatic systems | Elastomers, biomaterials | [47] |
|---|---|---|---|---|---|---|---|
| Biomimetic tentacle grippers | Marine organisms | Soft, slender structures, bending, suction | Underwater manipulation, exploration | Surgical tools, underwater exploration | Pneumatic or hydraulic systems | Elastomers, soft materials | [48] |

**8.4 Applications of bio-inspired soft grippers in bio-medical research**

In addition to tissue engineering and regenerative medicine, soft grippers have been studied for a range of other bio-medical applications. For example, they have been used in drug delivery, where they can be designed to release drugs at a specific location within the body. They have also been studied for use in surgical applications, such as minimally invasive surgery and neurosurgery, where their flexibility and precision make them ideal for delicate procedures. Overall, the development of soft grippers for bio-medical applications has been driven by the need for safe, adaptable, and precise devices that can handle delicate biological samples and perform a range of bio-medical tasks listed below.

**8.4.1 Surgical Assistants**: Soft robotic grippers can be used as surgical assistants to enhance the precision and dexterity of surgical procedures. They can manipulate and handle delicate tissues without causing damage, providing surgeons with improved control and reducing the risk of complications.

**8.4.2 Minimally Invasive Surgery:** Soft robotic grippers enable minimally invasive surgery, where small incisions are made instead of large open surgeries. These grippers can access confined spaces within the body and perform tasks such as tissue manipulation, suturing, and organ retraction with minimal trauma to surrounding tissues.

**8.4.3 Rehabilitation and Assistive Devices:** Soft grippers can be integrated into rehabilitation and assistive devices to aid individuals with impaired motor functions. They can assist in gripping and manipulating objects during rehabilitation exercises, enhancing the rehabilitation process and improving patient outcomes.

**8.4.4 Tissue Engineering: Artificial Skin/Nervous Tissue:** Soft robotics grippers have shown great promise in tissue engineering, particularly in the development of artificial skin and nervous tissue. These grippers, with their compliant and adaptable nature, can provide precise and gentle manipulation of delicate biological materials. Here, we explore the applications of soft robotics grippers in tissue engineering for artificial skin and nervous tissue.

**8.4.5 Bio-microfluidics Applications:** The application of microfluidic technology has brought about a paradigm shift in biomedical cell biology research conducted in vitro. Microfluidic channels are designed to replicate the confinement effects observed in physiological pathways, thereby providing a platform for studying cellular behavior in a more realistic and controlled environment. In recent studies, researchers have utilized microfluidic channels to fabricate soft robots that imitate biological networks.

**8.4.6 Endoscopic Procedures:** Soft robotic grippers integrated with endoscopic instruments allow for better maneuverability and control during endoscopic procedures. They can grasp and manipulate objects within the body, such as biopsy samples or foreign objects, aiding in diagnosis and treatment.

**8.4.7 Drug Delivery:** Soft grippers can be utilized in drug delivery systems to precisely and safely handle and transport drug-loaded carriers or micro/nanoparticles to targeted sites within the body. They enable controlled release of therapeutic agents, improving drug efficacy and reducing side effects.

**8.4.8 Prosthetics and Exoskeletons:** Soft robotic grippers find application in the development of prosthetic hands and exoskeletons. They offer improved dexterity and natural-like grasping capabilities, allowing users to perform intricate tasks with greater ease and functionality.

**8.4.9 Biopsy and Tissue Sampling:** Soft grippers are used in biopsy procedures to safely and accurately extract tissue samples for diagnostic purposes. Their compliant nature allows for gentle handling of delicate tissues, reducing patient discomfort and the risk of tissue damage.

**8.4.10 Microsurgery and Microscale Manipulation:** Soft robotic grippers excel in microscale applications, where delicate and precise manipulation is required. They can be employed in microsurgery, such as ophthalmic surgery or neurosurgery, to handle and position tiny structures with exceptional control.

The Table 8.2 summarizes various applications and examples of robotic grippers, highlighting their key features, bio-inspired working principles, actuation mechanisms, materials, sensing mechanisms, control strategies, morphology, and the physics of grasping. For each application, the corresponding gripper example is provided along with its distinctive attributes. The table underscores the innovation in robotics by elucidating how grippers are designed, influenced by nature, powered, controlled, and adapted for specific tasks, from gripping slippery tissues to enhancing hand rehabilitation and executing intricate endoscopic procedures.

TABLE 8.2 Comparison of different bio-inspired soft robotic grippers in terms of their applications.

| Application | References /Gripper Example | Key Features | Bio-inspired working Principles | Actuation Mechanism | Materials Selection | Sensing Mechanisms | Control | Morphology | Physics of grasping |
|---|---|---|---|---|---|---|---|---|---|
| Minimally Invasive Surgery | [49] | Gripping slippery and flexible tissues | Suction cup disc design inspired by Octopod Vulgaris | Hybrid Pneumatic and tendon-driven | Ecoflex, Nylon | Pneumatic (vacuum) sensing | Feedforward motion compensation control | Suction cup with flexible disk | Suction-based adhesion |
| Endoscopic Procedures | [50-52] | Bifunctional: one arm applies gripping force and retracts tissues, while the other arm | Dual arm endoscopic robot inspired by Norway Lobster | Hybrid tendon-sheath driven and SMA-based | Nitinol (NiTi) | Visual feedback control | Feedback control based on visual inputs | Dual arm endoscopic robot inspired by Norway Lobster | Friction-based |
| | | offers high maneuverability for cauterization of lifted tissues | | | | | | | |
| Rehabilitation and Assistive Devices | Soft Robotic Exo-Sheath for Hand Rehabilitation and Assistance [53] | Integrated soft hand exo-sheath with fabric-based EMG sensor for stroke and SCI patients' rehab and daily living aid | Bio-inspired fin-ray structure to enhance proprioception | DC Motor-based cable-driven | TPU (thermoplastic polyurethane) | Soft Fabric sEMG Sensor (Low resistance electrodes from silver fabric) | sEMG-based feedback control of finger flexion-extension (SVM classifier for patient intention) | Soft hand exo-sheath with fabric-based EMG sensor | Underactuation-based grasping |

| Tissue Engineering | Macroscale Muscle-Powered Robotic Gripper [54] | Macroscale muscle-powered robotic gripper | Muscle-inspired design | Pneumatic muscles | 3D-printed polymer | Optical encoder | Force feedback control | Multi-articulated gripper with pneumatic muscles | Muscle-like contraction and |
| --- | --- | --- | --- | --- | --- | --- | --- | --- | --- |
| Prosthetics and Exoskeletons | Multifinger Robotic Prosthesis [18] | Human-like grasping with force feedback from object | Human hand inspired tendon-driven gripper | DC Motor-based cable-driven | Aluminum | Strain gauge sensor for cable tension measurement | Closed-loop control of cable tension and stroke | Human-like fingered prosthetic hand | Force feedback and grip strength |

## 8.5 DESIGN PRINCIPLES OF BIO-INSPIRED SOFT GRIPPERS

### 8.5.1 Conventional vs. bio-inspired grippers

Conventional grippers and bio-inspired grippers have different design principles and approaches. conventional grippers are designed with a primary focus on functionality and efficiency, whereas bio-inspired grippers take inspiration from natural organisms and aim to replicate their unique gripping and manipulation abilities. They mimic the structures, mechanisms, or behaviors found in biological systems. The key differences between conventional grippers and bio-inspired grippers are presented in Table 8.3.

**TABLE 8.3** Key differences between conventional grippers and bio-inspired grippers

| Features\ Type of grippers | Conventional Grippers | Bio-Inspired Grippers |
|---|---|---|
| Design Focus | Functionality and efficiency | Biomimicry and adaptability |
| Structure | Rigid materials | Soft and flexible materials |
| Gripping Mechanisms | Standardized designs | Replication of biological gripping mechanisms |
| Actuation Systems | Electric motors, pneumatics, or hydraulics | Pneumatics, hydraulics, or tendon-driven systems |
| Flexibility | Limited flexibility and adaptability | High flexibility and adaptability |
| Handling | Suitable for general objects | Suitable for delicate objects and irregular shapes |
| Biomaterial Usage | Not specifically designed with biomaterials | Incorporation of biocompatible materials |
| Safety | May require additional safety measures | Designed for safe interaction with humans and delicate objects |

It is to be noted that the table provides a general comparison between conventional grippers and bio-inspired grippers and that specific gripper designs may vary within each category.

**8.5.2 Factors influencing soft gripper design for bio-medical applications**

While designing soft grippers for biomedical applications, several factors need to be considered to ensure their effectiveness and suitability for the intended purpose. Here are some key factors listed in Table 8.4 that influence soft gripper design in the context of bio-medical applications.

**TABLE 8.4** Key factors that influence soft gripper design in the context of bio-medical applications

| Features | Definition |
|---|---|
| Compliance and Softness | Soft grippers should be made from materials that are compliant and have a softness similar to biological tissues. This allows for gentle interaction with delicate structures and minimizes the risk of tissue damage or trauma |
| Biocompatibility | Soft grippers used in biomedical applications must be biocompatible to ensure they do not cause adverse reactions or harm to biological tissues. The materials used should be non-toxic and compatible with the physiological environment. |
| Flexibility and Adaptability | Soft grippers should possess flexibility and adaptability to conform to various shapes and sizes of biological structures or organs. This enables effective grasping and manipulation without causing undue stress or distortion to the tissues |
| Sensing and Feedback | Incorporating sensing capabilities into soft grippers can provide feedback on the grasping force, object properties, or tissue conditions. |

|  | This enables better control and ensures safe manipulation within the biomedical context |
|---|---|
| Sterilizability | Soft grippers used in medical settings should be capable of being sterilized to maintain aseptic conditions. Compatibility with common sterilization methods such as autoclaving, gamma irradiation, or chemical sterilization is crucial |
| Actuation Mechanism | The choice of actuation mechanism depends on the specific requirements of the biomedical application. Pneumatic, hydraulic, or shape memory alloy actuators may be used to achieve the desired motions and forces in the gripper |
| Size and Miniaturization | Some biomedical applications require soft grippers to be miniaturized to access confined spaces or perform minimally invasive procedures. Considerations for size, portability, and integration with other medical devices become essential in such cases |
| Controllability and Precision | Soft grippers should offer precise and controllable movements to perform delicate manipulations within the biomedical field. The design should allow for easy integration with control systems and provide fine-tuned control over the gripping forces and motions |
| Sterility and Cleanability | Soft grippers used in surgical or clinical settings should be designed for easy cleaning and sterilization. The gripper structure should minimize the risk of contamination and allow thorough cleaning between uses |
| Longevity and Durability | Soft grippers should be designed to withstand the rigors of biomedical applications and have good longevity. They should maintain their functional integrity over time and resist degradation from exposure to bodily fluids or sterilization procedures |

These factors need to be carefully considered and optimized during the design process to ensure that soft grippers meet the specific requirements of their intended biomedical applications.

### 8.5.3 Materials used in soft gripper fabrication for bio-medical applications

The choice of materials used in soft gripper fabrication for biomedical applications is critical to ensure biocompatibility, functionality, and durability. Some commonly used materials with their features for soft gripper fabrication in the biomedical field are listed in Table 2.5.

**TABLE 8.5** Materials with their features and possible application areas of soft gripper in the biomedical applications

| Materials | Features | Biomedical Applications |
|---|---|---|
| Elastomers | High elasticity, flexibility, biocompatibility, ease of fabrication | Surgical manipulators, wearable robots for human-motor interactions, cooperative robots for safety interactions, medical devices, prosthetics |
| Hydrogels | Water-swollen polymer networks, mimic soft tissue properties, high water content, biocompatibility | Tissue engineering scaffolds, drug delivery systems, wound dressings, organ-on-a-chip devices |
| Thermoplastic Elastomers (TPEs) | Versatile, flexibility, elasticity, moldability, 3D printability, wide range of durometer options | Prosthetics, orthotics, wearable sensors, rehabilitation devices |
| Shape Memory Polymers (SMPs) | Deformable and recoverable shape, stimuli-responsive, adaptive designs | Minimally invasive surgical tools, stents, catheters, tissue scaffolds |
| Biomaterials | Biodegradable, biocompatible, temporary implants, gradual degradation over time | Bioresorbable surgical implants, drug delivery systems, tissue engineering constructs |

| Conductive Materials | Electrical conductivity, integration with sensing technologies, conductive elastomers, composites with nanoparticles | Bioelectronic devices, sensors for physiological monitoring, neural interfaces |
|---|---|---|
| Adhesives | Secure bonding, biocompatible, assembly and attachment of soft gripper components | Assembly of biomedical devices, attachment of sensors or actuators to biological tissues |

## 8.6 SOFT GRIPPER FABRICATION TECHNIQUES FOR BIO-MEDICAL APPLICATIONS

Fabrication techniques for biomedical applications encompass a range of methods used to create functional and tailored devices, structures, or materials with specific properties for use in the field of medicine. These techniques play a crucial role in developing biomedical devices, tissue engineering scaffolds, drug delivery systems, and other healthcare-related applications. Here is an overview of commonly employed fabrication techniques in the biomedical field:

**8.6.1 Additive Manufacturing (3D Printing):** 3D printing enables the layer-by-layer fabrication of complex geometries using various materials, such as polymers, metals, ceramics, or hydrogels. It allows for precise customization, rapid prototyping, and the incorporation of intricate features, making it suitable for creating patient-specific implants, tissue scaffolds, drug-loaded microparticles, and anatomical models.

**8.6.2 Electrospinning**: Electrospinning involves the use of an electric field to create nanofibers from polymer solutions or melts. This technique can produce fibrous structures with high surface area-to-volume ratios, mimicking the extracellular matrix (ECM) for tissue engineering applications. Electrospinning is commonly used to create scaffolds for wound healing, tissue regeneration, and drug delivery.

**8.6.3 Microfluidics:** Microfluidics utilizes microscale channels, chambers, and valves to manipulate small volumes of fluids. It enables precise control over fluid flow, mixing, and biochemical reactions. Microfluidic devices are employed for applications like organ-on-a-chip systems, drug screening, point-of-care diagnostics, and lab-on-a-chip platforms.

**8.6.4 Bioprinting**: Bioprinting is a specialized form of 3D printing that focuses on fabricating living tissues or organs. It involves the deposition of bioinks containing living cells, biomaterials, and growth factors to create complex, functional tissue structures. Bioprinting holds promise for tissue and organ transplantation, disease modeling, and drug testing.

**8.6.5 Soft Lithography:** Soft lithography uses elastomeric materials, typically polydimethylsiloxane (PDMS), to create microstructures or microfluidic devices. It involves casting a master mold, typically created through photolithography, and replicating the pattern onto PDMS. Soft lithography techniques are employed in creating microfluidic devices, cell culture platforms, and biomedical sensors.

**8.7.6 Electrochemical Deposition:** Electrochemical deposition, also known as electrodeposition, involves the electrodeposition of materials onto a substrate using an electric current. This technique allows for precise control over the coating thickness and the incorporation of various materials, such as metals or hydroxyapatite, for applications such as implants or biomedical coatings.

**8.7.7 Microfabrication:** Microfabrication techniques, such as photolithography, thin film deposition, and etching, are utilized to create microscale structures and devices. They are essential for manufacturing microelectromechanical systems (MEMS), microfluidic devices, biosensors, and lab-on-a-chip platforms.

**8.7.8 Biomimetic Self-Assembly**: Biomimetic self-assembly techniques utilize the self-organization and self-assembly properties of biological molecules, such as proteins or peptides, to create functional structures. This approach is used to fabricate biomimetic materials, drug delivery carriers, and tissue-engineered constructs.

**8.7.9 Melt Processing:** Melt processing involves the melting and subsequent solidification of thermoplastic materials to create desired shapes. Techniques such as injection molding, hot pressing, or hot extrusion are commonly employed for mass production of biomedical devices, implants, or drug delivery systems.

These fabrication techniques offer versatile and precise approaches to create structures, devices, or materials tailored for biomedical applications. The choice of technique depends on factors such as the desired properties, complexity, scalability, and compatibility with biological systems.

When it comes to fabricating biomedical devices, structures, or materials, various techniques are available, each with its own advantages and limitations. A comparison of different fabrication techniques commonly used in biomedical applications is presented in Table 8.6.

**TABLE 8.6** Comparison of different fabrication techniques for bio-medical applications

| Fabrication Technique | Advantages | Limitations |
|---|---|---|
| Additive Manufacturing | Customizable, complex geometries, rapid prototyping, diverse material options | Limited resolution, slower printing speeds for large objects, material properties may differ from bulk counterparts |
| Electrospinning | High surface area, nanoscale fibers, tunable porosity, mimics extracellular matrix | Limited control over fiber alignment, challenging to scale up |
| | | production, difficulties with cell viability |
| Microfluidics | Precise fluid control, miniaturization, high-throughput capabilities, compatibility with lab-on-a-chip | Limited scalability, challenges in integrating multiple functionalities, fabrication complexity |
| Bioprinting | Enables 3D cell encapsulation, tissue-specific architecture, potential for personalized medicine | Limited printing resolution, challenges in vascularization and long-term functionality of printed constructs |
| Soft Lithography | High-resolution patterning, flexible materials, compatibility with microfluidics | Limited scalability, challenges with certain material properties, complex fabrication processes |
| Electrochemical Deposition | Precise coating thickness control, compatibility with various materials | Limited complexity in structure, challenges with complex geometries and internal structures |
| Solvent Casting and Particulate Leaching | Porous scaffold fabrication, control over pore size and distribution | Limited control over pore interconnectivity, challenges in maintaining mechanical integrity and cell infiltration |
| Microfabrication | High precision, miniaturization, compatibility with MEMS devices | Limited scalability, complex fabrication processes, challenges with certain materials and geometries |
| Biomimetic Self-Assembly | Biomimetic structures, self-organization properties, | Limited control over assembly process, challenges in scaling up |

|  | potential for functional materials | production and maintaining stability |
|---|---|---|
| Melt Processing | Mass production capability, compatibility with various thermoplastic materials | Limited resolution, challenges with complex geometries and material degradation at high temperatures |

## 8.7 PERFORMANCE EVALUATION OF SOFT GRIPPERS FOR BIO-MEDICAL APPLICATIONS

There are several factors which influence the performance of the grippers and to measure the performance several metrics could be used in bio-medical applications.

Each metric provides specific insights into the gripper's capabilities, ensuring that it meets the requirements for a particular application. These metrics are crucial for evaluating the performance of grippers in bio-medical applications and they are listed in the Table 8.7.

**TABLE 8.7** Metrics for evaluating gripper performance in bio-medical applications

| Metric | Description |
|---|---|
| Grasping Force | Measures the gripping strength of the gripper |
| Dexterity | Evaluates the gripper's ability to manipulate objects with precision and control |
| Compliance and Gentle Handling | Assesses the gripper's ability to interact gently with delicate tissues or objects |
| Adaptability to Object Shape/Size | Measures the gripper's ability to adapt to various object geometries |

| Biocompatibility | Evaluates the gripper's compatibility with biological tissues or fluids |
|---|---|
| Controllability and Accuracy | Assesses the gripper's precision, control accuracy, and repeatability |
| Sterilizability | Determines the gripper's compatibility with sterilization methods |

## 8.8 CHALLENGES AND FUTURE DIRECTIONS

### 8.8.1 Challenges faced in soft gripper development and adoption for bio-medical applications

The development and adoption of soft grippers for bio-medical applications come with several challenges that need to be addressed. These challenges can impact various aspects of gripper design, fabrication, functionality, and integration into the biomedical field. We listed below are some key challenges faced in soft gripper development and adoption for bio-medical applications:

- **Biocompatibility:** Ensuring the biocompatibility of soft grippers is crucial when they come into direct contact with biological tissues or fluids. The selection of materials and surface treatments must be carefully considered to minimize adverse reactions, inflammation, or cytotoxicity.

- **Robustness and Durability:** Soft grippers need to be durable and robust enough to withstand the rigors of biomedical applications. The materials used should maintain their integrity over time, withstand repetitive use, and resist degradation from exposure to bodily fluids, sterilization procedures, or mechanical stresses.

- **Control and Sensing:** Achieving precise control over soft grippers can be challenging due to their compliance and flexibility. Integrating accurate sensing mechanisms to provide feedback on grip force, object properties, or tissue conditions is essential for safe and effective manipulation.

- **Grasping Complexity:** Soft grippers may face challenges when grasping complex objects with irregular shapes, varying sizes, or fragile structures. Designing grippers with adaptive mechanisms or surface features to enhance grip stability and adaptability is crucial for reliable grasping performance.

- **Scalability and Manufacturing:** Scaling up the production of soft grippers while maintaining consistent performance can be challenging. Fabrication techniques must be scalable and compatible with high-volume manufacturing processes to meet the demands of biomedical applications.

- **Sterilization:** Soft grippers used in biomedical settings require proper sterilization to ensure aseptic conditions. However, some materials or fabrication techniques may be sensitive to certain sterilization methods, limiting the options for maintaining sterility.

- **Integration with Other Systems:** Integrating soft grippers into existing biomedical systems, such as robotic platforms or surgical tools, can present compatibility and integration challenges. Ensuring seamless integration, interface compatibility, and communication protocols is essential for effective use in clinical or surgical environments.

- **Regulatory Compliance:** Soft grippers developed for biomedical applications need to comply with regulatory standards and undergo rigorous testing and validation processes. Meeting safety, efficacy, and quality standards set by regulatory bodies can be time-consuming and resource-intensive.

- **User Acceptance and Adoption:** The successful adoption of soft grippers in bio-medical applications depends on user acceptance, training, and familiarity. Overcoming any resistance to new technologies or changing conventional practices is crucial for their widespread adoption.

- **Cost Considerations:** The cost of developing and implementing soft grippers for bio-medical applications can be a significant challenge. Balancing the cost of materials, fabrication, and scalability while ensuring performance and quality can impact the feasibility and commercial viability of soft gripper solution.

Addressing these challenges through interdisciplinary collaborations, continuous research, and technological advancements will pave the way for the successful development and adoption of soft grippers in bio-medical applications.

### 8.8.2 Opportunities for improvement and innovation

Soft gripper research for bio-medical applications presents numerous opportunities for improvement and innovation. Advancements in materials, design, fabrication techniques, and integration can enhance the performance, functionality, and applicability of soft grippers in the biomedical field. Here are some key opportunities for improvement and innovation in soft gripper research for bio-medical applications:

- **Enhanced Biocompatibility:** Developing novel biocompatible materials and surface treatments that minimize adverse reactions, promote tissue integration, and reduce cytotoxicity will improve the biocompatibility of soft grippers for safe and effective interaction with biological systems.

- **Customizability and Personalization:** Advancements in 3D printing and bioprinting technologies allow for the customization and personalization of soft grippers. Tailoring grippers to patient-specific anatomies or specific biomedical tasks can improve their performance and patient outcomes.

- **Sensing and Feedback Integration:** Integrating advanced sensing technologies, such as force sensors, tactile sensors, or biosensors, into soft grippers will enable real-time feedback and improved control during grasping and manipulation tasks. This integration can enhance safety, precision, and adaptability.

- **Biomimetic Design:** Drawing inspiration from natural organisms and structures can lead to innovative gripper designs. Mimicking biological systems, such as the human hand or specific animal appendages, can result in soft grippers with improved dexterity, adaptability, and functionality.

- **Smart and Active Materials:** Researching and developing smart materials, such as shape memory polymers, electroactive polymers, or stimuli-responsive hydrogels, can enable soft grippers with active functionalities, such as self-healing, shape-changing, or stimuli-triggered gripping and releasing.

- **Multi-modal Actuation:** Exploring hybrid actuation mechanisms combining pneumatic, hydraulic, electrical, or magnetic actuation can provide soft grippers with versatile and multi-modal capabilities, expanding their range of applications and enabling complex tasks.

- **Autonomous and Intelligent Control:** Advancements in artificial intelligence (AI) and machine learning (ML) can be applied to soft gripper control, enabling autonomous grasping, object recognition, and adaptive manipulation. Intelligent control algorithms can improve efficiency, adaptability, and overall performance.

- **Miniaturization and Integration with Microsystems:** Downsizing soft grippers to microscale or integrating them with microsystems, such as microfluidics or microelectromechanical systems (MEMS), can open up new possibilities for minimally invasive surgical tools, lab-on-a-chip devices, or implantable biomedical devices.

- **Bioinspired Soft Robotics:** Leveraging principles from nature, such as self-healing, self-regulation, or hierarchical structures, can inspire the development of bioinspired soft grippers with enhanced functionalities, adaptability, and resilience.

- **Human-Machine Interaction and Collaboration:** Designing soft grippers for seamless integration and interaction with humans, robotic systems, or surgical tools can enable safe and effective human-machine collaboration, expanding the possibilities for cooperative biomedical procedures.

Exploring these opportunities through interdisciplinary collaborations, research partnerships, and technological advancements will drive innovation in soft gripper research for bio-medical applications, leading to safer, more effective, and transformative solutions in the field of medicine.

## CONCLUSION

The chapter provides a comprehensive overview of the design principles, fabrication techniques, challenges, and opportunities in the field. It highlights the significance of these soft grippers in enabling safer interactions, precise manipulations, and innovative solutions in various bio-medical applications, including surgery, tissue engineering, and diagnostics. In summary, bio-inspired soft grippers hold great promise for bio-medical applications due to their unique characteristics such as compliance, adaptability, and biocompatibility. They offer advantages over conventional rigid grippers, enabling safer interactions with biological tissues and facilitating complex tasks in medical settings. The development and adoption of soft grippers in bio-medical applications face challenges related to biocompatibility, robustness, control, scalability, and integration. However, there are significant opportunities for improvement and innovation, including enhanced biocompatibility, customizability, sensing integration, biomimetic design, smart materials, multi-modal actuation, and intelligent control. Addressing these challenges and leveraging these opportunities will drive advancements in soft gripper research, leading to transformative solutions in the field of bio-medical applications, such as surgical manipulators, wearable robots, tissue engineering, and minimally invasive procedures.

Future directions for soft bio-inspired grippers in biomedical applications involve interdisciplinary research collaborations and advancements in material science, robotics, and biomedical engineering. This includes developing novel materials with improved mechanical properties, enhanced biocompatibility, and integrated sensing functionalities. Moreover, the integration of advanced control algorithms and artificial intelligence techniques can enable more sophisticated and autonomous gripper behaviors. Furthermore, exploring new application areas within biomedicine, such as minimally invasive surgery, targeted drug delivery, and tissue engineering, holds great promise for soft bio-inspired grippers. By addressing specific challenges in these domains, such as delicate tissue handling, precise drug administration, and tissue manipulation, soft bio-inspired grippers can significantly contribute to advancements in healthcare and biotechnology.

In conclusion, soft bio-inspired grippers have demonstrated their potential in biomedical applications, offering unique capabilities for delicate object manipulation and interaction with living tissues. Overcoming challenges related to dexterity, adaptability, sensing, and biocompatibility will pave the way for their wider adoption and integration into various biomedical procedures. With continued research and innovation, soft bio-inspired grippers hold tremendous promise for revolutionizing biomedical technologies and improving patient outcomes in the future.

**Authors Biography:**

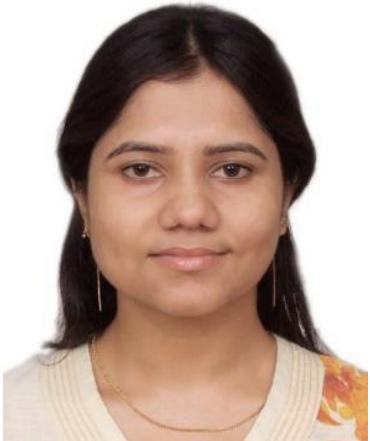

**Rekha Raja** obtained her Bachelor's degree from Jalpaiguri Govt. Engineering College, India, in 2007, Master's degree from Bengal Engineering and Science University, India, in 2009 and PhD degree from the Indian Institute of Technology Kanpur, India in 2016 . From 2016 to 2018 she worked as a scientist, at Innovation Lab, Tata Consultancy Services, Noida, India. She did her postdoctoral research at the University of California Davis, USA from 2018 to 2019 and at Wageningen University, Netherlands from 2019 to 2023. Currently, she is an Assistant Professor at the department of Artificial Intelligence at the Radboud University, Nijmegen, Netherlands. Her research interests include Robotics, AI, Robot motion planning, computer vision, machine learning, deep learning etc.

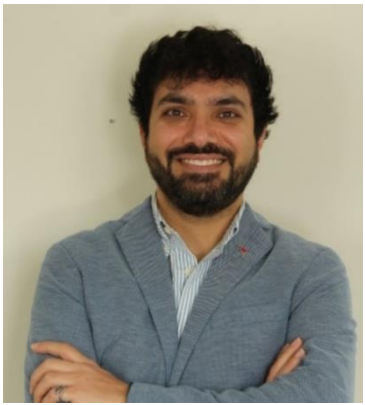

**Ali Leylavi Shoushtari** received his Ph.D. in Biorobotics from Scuola Superiore Sant'Anna, Pisa, MSc in Mechatronics from South Tehran Branch, Azad University and Bsc in Robotics Engineering from Shahrood University of Technology in Iran. He pursued research on the design, fabrication, and modeling of soft origami actuators at the Center for Micro-BioRobotics (CMBR), IIT as a postdoctoral researcher. In 2019 he joined Wageningen University & Research where he worked on advanced soft grippers for delicate crop handling. Currently he is senior scientist at Wageningen Robotics working on Research and development of soft robotic technologies. His research interest is Robotics manipulation and control, design and fabrication of soft robots and their application in agriculture and food domain.